\documentclass[letterpaper]{article} 
\ifdefined\aaaianonymous
    \usepackage[submission]{aaai2027}  
\else
    \usepackage{aaai2027}              
\fi
\usepackage[hyphens]{url}  
\usepackage{graphicx} 
\usepackage{natbib}  
\usepackage{caption} 
\usepackage{amsmath}
\usepackage{amssymb}
\usepackage{booktabs}
\usepackage{makecell}
\title{When Counterbalancing Hides the Bias:\\Access-Conditioned Position Lock in Forced-Choice LLM Evaluation}

\author{
    Hong-In Won\textsuperscript{1,2}\thanks{Corresponding authors: Hong-In Won (\texttt{luvhayym@kitech.re.kr}), Hyoseop Kim (\texttt{hyoseop1231@kitech.re.kr}).},\quad
    Jinseok Jang\textsuperscript{3},\quad
    Hyoseop Kim\textsuperscript{1}\footnotemark[1]
}
\affiliations{
    \textsuperscript{1}Korea Institute of Industrial Technology (KITECH), Incheon, Republic of Korea\\
    \textsuperscript{2}National Research Council of Science and Technology (NST), Sejong, Republic of Korea\\
    \textsuperscript{3}Korea Institute of Industrial Technology (KITECH), Daegu, Republic of Korea\\
    \smallskip
    ORCID: Hong-In Won 0000-0003-1609-8447, Jinseok Jang 0000-0002-6801-9109, Hyoseop Kim 0009-0005-7472-4073
}

\begin{document}
\maketitle

\begin{abstract}
Forced-choice probes with counterbalanced orientations are a standard tool for measuring language-model ``value dispositions,'' and a concentration/extremity index over repeated draws is read as how sharply a model commits. We show this estimator is not identifiable at its low end: counterbalancing, meant to remove position bias, instead maps a position-lock (a model returning the same answer \emph{letter} regardless of content) onto the same near-0.5 signature as genuine neutrality, so ``softness'' and ``non-engagement'' cannot be distinguished from a content-independent letter-bias. Across nine models the fraction of position-locked items tracks the index almost perfectly ($r = -0.986$) --- a structural consequence, not a finding: the informative quantity is the residual from that bound, the commitment a model shows on the items it does engage. The three models the index reads as soft are the most position-locked, and the lock resolves under access configurations that permit reasoning (as-deployed CLI $\to$ raw API $\to$ reasoning-enabled), moving the index e.g.\ $0.06\to0.63$ and $0.34\to0.66\to0.64$ while lock collapses (Opus, \texttt{claude-opus-4-8}: $0.61\to0.39\to0.22$); this establishes the low readings as an identifiability failure of the instrument, not a disposition. The reasoning-enabled reading is not a ``true'' value either; the point is that the index alone cannot identify commitment at its low end. Access configuration (deployment client, reasoning on/off) is one generator of this failure, shown on two access paths (an Anthropic subscription CLI and a constrained-budget DeepSeek client), and in a frozen benchmark it is confounded with provider. We contribute a position-lock diagnostic that must accompany any concentration reading, and show that a concentration-blind audit risks reporting neutrality where a reasoning-permitting condition yields concentrated choices. The direction-flip component, largely robust to the artifact, still identifies genuine cross-model disagreement.
\end{abstract}

\section{1. Introduction}

Language models are now routinely compared across providers for their ``value dispositions'' --- the side a model takes when a prompt poses a value trade-off with no rule-governed answer. A common instrument is a counterbalanced forced-choice probe: pose a binary dilemma, present each option under both answer labels, repeat the draw, and summarize the distribution with a concentration index. A low index reads as soft or non-committal, a high one as sharp commitment, and these readings feed alignment audits and model-selection decisions.

\begin{figure}[t]
\centering
\includegraphics[width=\columnwidth]{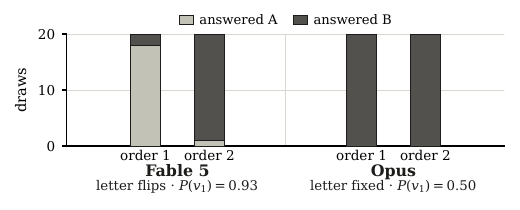}
\caption{Counterbalancing averages a letter-lock into apparent neutrality. One dilemma (D04),
20 draws under each label ordering. Fable 5's answer letter changes when the labels swap, so its
value choice is held at $P(v_1)=0.93$. Opus returns B on 20 of 20 under \emph{both} orderings, so
the value it selects flips with the labels and counterbalancing averages it to exactly 0.50 --- the
same reading a genuinely balanced model would produce. The concentration index sees only that 0.50;
the per-item letter-invariance rate is what tells the two apart.}
\label{fig:mechanism}
\end{figure}

\textbf{Our central result is that this instrument is not identifiable at its low end.} Counterbalancing removes the position bias it was designed for, but also does something the concentration reading does not account for: it maps a \textbf{position-lock} (a model that returns the same answer \emph{letter} irrespective of the content behind the labels) onto exactly the near-0.5 signature it produces for a genuinely balanced disposition. Consider a student who never reads the question and always marks the first option. An examiner who swaps the option order on half the papers, precisely to remove position bias, records that student at exactly 50\% and reads it as balanced judgement between the two positions. Counterbalancing did not remove the bias; it averaged it out of sight (Figure~\ref{fig:mechanism}). At the soft end, ``softness'' is therefore indistinguishable from a \textbf{content-independent letter-bias}, and the index alone cannot identify which a low reading reflects.

\textbf{This is not a hypothetical degeneracy.} Across our nine models, the fraction of dilemmas answered with the same letter under both orientations (a direct position-lock measurement) correlates with the concentration index at $r = -0.986$ (a descriptive tightness measure; the two quantities are defined on the same items and structurally constrained, so this is not a significance test --- see \S4). At the low end ``softness'' is very largely a re-encoding of position-lock: the three models read as soft (Opus 0.34, Sonnet 0.48, DeepSeek 0.06) are the three most position-locked. Sonnet is the weak case: its 9 of 18 does not separate from the chance floor a genuinely indifferent model produces (Appendix~C), and we do not count it as established. The lock is not fixed. As the access path is made progressively more reasoning-permitting (as-deployed subscription CLI $\to$ raw provider API $\to$ reasoning-enabled), position-lock collapses monotonically (Opus $0.61\to0.39\to0.22$) and the index jumps off its as-deployed low (Opus $0.34\to0.66\to0.64$; DeepSeek $0.06\to0.63$). That progression pins the low readings to the \emph{access path}, not the disposition: what the standard instrument reports as a value trait is, for these three models, an identifiability failure.

\paragraph{What we do and do not claim.} We do \textbf{not} claim the reasoning-enabled reading is the model's ``true'' value. There is no single ground truth to recover: what a model chooses is measurement-conditioned, and the reasoning-enabled path is not a privileged oracle. What it establishes is narrower and sufficient --- a condition under which the instrument measures \textbf{value-engagement rather than letter-bias}, demonstrating that the low reading was uninterpretable, not that the high reading is canonical. Position-lock is therefore not \emph{another valid value} held under one condition but an \textbf{invalid reading}, the instrument measuring position rather than value. This asymmetry --- some readings invalid, the rest condition-dependent but valid --- is the whole claim, and it differs from both ``we can measure a model's true values'' and ``the harness changes a model's values.''

\paragraph{The positive residual.} Not everything the instrument reports collapses under this diagnosis. The component that turns on \emph{direction} --- whether two models fall on opposite sides of the choice, a genuine value disagreement --- is largely robust to the position-lock artifact, because a position-locked model sits at the boundary and cannot direction-disagree with anyone. Once the non-identifiable low-end magnitude is set aside, the direction-flip structure still isolates real cross-model disagreement, and a subset of cross-family disagreements survives a strict near-boundary robustness test. Identifying \emph{what remains identifiable} is as much the contribution as diagnosing what does not.

\paragraph{Contributions.} First, a position-lock identifiability diagnostic (\S3, \S4): we show the counterbalanced concentration index is not identifiable at its low end, quantify the coupling (position-lock fraction $\leftrightarrow$ index, $r = -0.986$), and prescribe that a per-model position-lock fraction accompany \emph{every} concentration reading --- a low index is uninterpretable until position-lock is excluded. Second, a direction-flip decomposition that isolates what is still identifiable (\S4): we split an apparent cross-model distance into a direction-flip component (a disagreement largely robust to the artifact) and a same-side magnitude residual that is not identifiable at the low end, and show the field's naive single-draw distance mixes the two. Third, access configuration as one generator (\S5): the deployment client and reasoning-on/off setting are one route by which position-lock is induced (shown on the two access paths named in the abstract), invisible at the output, enough to flip a model between ``value-neutral'' and ``value-committed,'' and confounded with provider in a frozen benchmark.

This is a methods paper: its contribution does not ride on whether models prove individuated or convergent, only on the diagnostic that makes the low-end reading interpretable at all.

\section{2. Related Work}

\textbf{The work this paper argues against is single-draw counterbalanced cross-model value comparison.} A growing body of work compares language models by their choices on moral or value dilemmas and reads the cross-model differences (or convergences) as facts about the models' values: moral-belief benchmarks (MoralChoice~\citep{scherrer2023moralchoice}), value inventories (ValueBench~\citep{ren2024valuebench}; Value Portrait~\citep{han2025valueportrait}), and dilemma corpora scored by chosen-minus-neglected values (DailyDilemmas~\citep{chiu2024dailydilemmas}), some reporting cross-model \emph{convergence} across ten frontier models \citep{zhang2026moralcomposition, huang2026knowing}. These vary in how they sample: some record one response per item, others (e.g.\ MoralChoice) sample repeatedly and report a decidedness split. We target no published pipeline, but an inferential rule assembled from practices in use: counterbalance labels, summarize the resulting choices with a concentration index, and read that index or its cross-model differences as evidence about model values. At the low end, this rule is non-identifiable because counterbalancing makes a position-locked model and a genuinely neutral model yield the same concentration. Counterbalancing is the standard debiasing fix \citep{dominguez2023survey} and our measurement \emph{is} such a correction; the failure we identify is second-order, entering in the summary through which that correction is read. Closest to our aim, \citet{kriegmair2026individuality} separate genuine individuality from a response-style confound, but partition \emph{within-model} variance over continuous ratings, whereas we identify a \emph{non-identifiability} of a counterbalanced concentration index and decompose a \emph{between-model} distance. Nearest on the access axis, value-elicitation work varies the elicitation and finds it changes the values read out (\citet{rottger2024political}; ``Mind the Gap''~\citep{mahajan2026mindthegap}), and a forced-choice study finds constraining a prompt to multiple choice \emph{induces} compliance \citep{chen2026choices}; agent-comparison work argues results are uninterpretable without disclosing the harness \citep{zhang2026harness, sclar2024formatspread, yao2026harnessbench}. Our contribution is the intersection none occupy: (i) the identification result, that the counterbalanced concentration index cannot separate a position-lock from genuine neutrality at its low end; (ii) that \textbf{access configuration} is \emph{one generator} of that lock, common and invisible in real deployment; and (iii) a white-box control tracing the induced compliance to the deployment system prompt (\S5). An extended treatment --- within-model concentration/decidedness measures \citep{scherrer2023moralchoice, atil2024nondeterminism, moore2024consistent, lee2024inertia, hadarshoval2024alignment}, persona/identity drift \citep{li2024instability, kim2024identitydrift}, and provider-layer steering \citep{sharma2023sycophancy, denison2024subterfuge} --- is in Appendix~A.

\section{3. Measurement Protocol}

\textbf{Stimulus and counterbalanced base measurement.} Each item is a forced binary choice between two options instantiating a value trade-off, constructed so that (a) no legal, safety, or contractual rule forces one option (so the choice is genuinely a value choice); (b) the actor is in-scope to decide, removing ``not my call'' refusals; and (c) the two options are surface-symmetric in length and framing. We use 18 dilemmas (an eight-item core set of workplace trade-offs and a ten-item extended set probing Schwartz value axes and Moral Foundations dimensions); all are benign, synthetic, and free of personal data; the item-to-axis mapping is given in Appendix~H. Items are posed in Korean and answered by a forced single letter; we test the main ordering's robustness to prompt language by re-running the full protocol with English-translated items (Appendix~F). For each model and item we collect $N=40$ independent responses under a no-persona configuration. To remove position bias, option labels are counterbalanced across draws (the designated side $v_1$ appears as ``A'' in half and ``B'' in the other half), and we record $P(v_1)$. Counterbalancing is also what makes position-lock \emph{observable in principle}: because the designated side appears under both labels, a content-tracking model flips its letter when labels swap and a position-locked one does not. This letter-invariance check is the raw material for the diagnostic below --- yet the same counterbalancing, once summarized by the concentration index, averages the lock into a near-0.5 signature indistinguishable from neutrality, so the diagnostic must be computed and reported separately.

\textbf{Concentration index and the position-lock diagnostic.} We summarize how concentrated a model's counterbalanced choices are with an \textbf{extremity} index,
\[
\text{extremity} = \text{mean over items of } |P(v_1) - 0.5| \times 2,
\]
which is 1 when a model is fully concentrated on every item ($P \in \{0,1\}$) and 0 when the choices sit at $P = 0.5$ everywhere. We name it by what it measures (the concentration of counterbalanced choices) not ``determinism,'' because a low value does not identify low commitment. A reading of $P(v_1)\approx 0.5$ has three causes the index conflates (Figure~\ref{fig:mechanism}): a genuine balanced disposition, a graded lukewarm leaning, and \textbf{position-lock}, where the model fixes on an answer letter and counterbalancing averages it to 0.5. Only the third is a measurement artifact, and it is continuous: across our nine models the fraction of dilemmas answered with the \emph{same letter in both orientations} correlates with extremity at $r = -0.986$. We therefore \textbf{report a per-model position-lock fraction alongside every extremity value} and treat a low extremity as \emph{uninterpretable until position-lock is excluded}. We do not stop at correlation: re-measurement under reasoning-permitting access paths (\S4, \S5) shows access-conditioned instability with a concurrent drop in lock, and the white-box control shows an agent-style prompt is \emph{sufficient} to move compliance and profile. Re-measurement establishes that a low reading was \emph{invalid} (the instrument measuring letter-position rather than value); it does not establish the reasoning-enabled reading as the model's \emph{true} value.

\textbf{Profile distance and controls.} For two models the per-item distance is $|\Delta P(v_1)|$ and the profile distance its mean over the 18 items --- the quantity a cross-model comparison implicitly reports, which \S4 shows can be inflated purely by position-lock at the low end. To measure whether a \emph{declared} value-persona shifts a disposition we use a \textbf{neutral-persona control} (a value-irrelevant preamble) rather than the bare base, since any preamble perturbs the base distribution; the persona/resistance axes are deferred to Appendix~D. Because items are forced binary choices, scoring is a single-letter parse de-mapped to the designated side, requiring no blind or model-based judging.

\textbf{Statistics and reporting convention.} Interval estimates are \textbf{95\% bootstrap confidence intervals} (10,000 resamples, seed 42) for per-pair genuine shares and profile shifts; the one exception is the position-lock$\leftrightarrow$extremity correlation, for which we report a \textbf{Fisher-z interval} (\S4) because a percentile bootstrap saturates the $[-1,1]$ boundary at $n=9$. It is a \emph{selected-model sensitivity interval}, not a population one (Appendix~I). With $N=40$ the standard error at $P=0.5$ is $\approx0.08$, half-width $\approx0.16$; the $\pm0.1$ dead-band of \S4 is deliberately looser, and \S4 reports $\pm0.16$ beside it. \textbf{Provenance and data integrity} procedures (raw-output storage; a datasheet incident log recording two mid-run binary-swap failures and a parser-contamination incident that fabricated a spurious soft reading, distinct from position-lock) are in Appendix~B.

\section{4. The Concentration Index Is Not Identifiable at Its Low End}

\textbf{The identifiability failure, stated and measured.} Take a reading of $P(v_1)\approx 0.5$ under counterbalanced measurement. Three generators produce this signature (\S3, Figure~\ref{fig:mechanism}); one of them, position-lock, is an artifact. The index cannot distinguish them: it is \emph{not identifiable} at its low end. That this is not a corner case is shown by the position-lock fraction, which across the nine models tracks extremity almost perfectly (Figure~\ref{fig:scatter}; per-model values in Appendix~C, Table~C2): the Pearson correlation is $r = -0.986$ (Fisher-z interval $[-0.997, -0.932]$, which respects the $[-1,1]$ boundary a percentile bootstrap saturates at $n=9$). We read this \emph{descriptively}, not as a hypothesis test: extremity and position-lock are defined on the same 18 items and structurally related, a position-locked item sits at $P\approx0.5$ and contributes $\approx 0$ to extremity, so extremity $\le 1 - \text{lock\_fraction}$ is an approximate upper bound, the observed points other than DeepSeek sit at 82--100\% of it (Table~C2), and $r$ is pushed toward $-1$ mechanically. Here $r=0$ is not the relevant null, so testing against it would mislead. The bound is \emph{loose}, not an identity: sampling noise and partial locking leave slack, and DeepSeek violates it outright ($0.06 > 1-1.00$). The informative quantity is therefore not the correlation but the \emph{residual} from the bound --- the engaged-subset commitment Appendix~C's within-subset extremity reports directly. The paper's empirical weight rests on the re-measurement result and the white-box control (\S5), not on this correlation. The chance floor is \emph{not} one half but $0.339$ (Appendix~C): DeepSeek's 18/18 and Opus's 11/18 ($p\!\approx\!0.017$) separate from it, \textbf{Sonnet's 9/18 does not} ($p\!\approx\!0.12$) and is reported as undetermined.

\begin{figure}[t]
\centering
\includegraphics[width=0.8\linewidth]{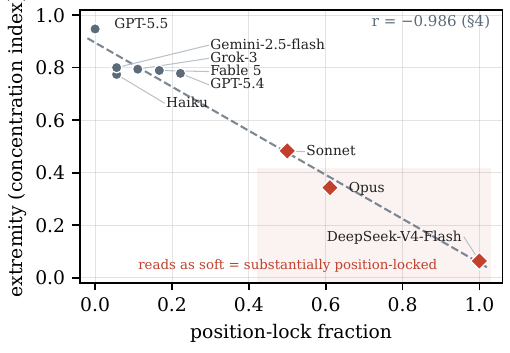}
\caption{Extremity (vertical) against position-lock fraction (horizontal), nine models ($r = -0.986$, a descriptive tightness measure; the two axes are structurally constrained, \S4, not a significance test). ``Softness'' (low extremity) is almost entirely explained by position-lock: the three models the index reads as soft --- Opus, Sonnet, DeepSeek (diamonds) --- are also the three most position-locked ones. The shaded corner marks the two whose lock is established; Sonnet sits just above it, its 9 of 18 not separating from the chance floor (Appendix~C). The relationship is the artifact, not a disposition.}
\label{fig:scatter}
\end{figure}

\paragraph{Note on models.} We study nine models across five provider families: Anthropic \{Opus (\texttt{claude-opus-4-8}), Sonnet (\texttt{claude-sonnet-5}), Haiku (\texttt{claude-haiku-4-5}), Fable 5 (\texttt{claude-fable-5})\}, OpenAI \{GPT-5.5, GPT-5.4\}, Google \{Gemini-2.5-flash\}, DeepSeek \{DeepSeek-V4-Flash\}, and xAI \{Grok-3\}; Appendix~G pins each snapshot and collection date, and Appendix~K asks whether the lock persists on later models. Short names denote exactly the pinned identifiers above and no later model of the same family; where a figure could span generations we give the identifier. The Anthropic extremities are measured as-deployed through the subscription CLI, one generator of position-lock (\S5).

\textbf{Re-measurement confirms an artifact, not a disposition.} Correlation alone leaves open whether the three soft models are position-locked \emph{because} of the access path or merely happen to be both. We resolve this by re-measuring them under access paths that progressively permit reasoning --- as-deployed subscription CLI (bare-letter, default budget), raw provider API, and a reasoning-enabled configuration (adaptive thinking at maximum effort) --- all at $N=40$. As the path admits more reasoning, the position-lock collapses stepwise and only for the position-locked models (Table~\ref{tab:progression}; also plotted in Appendix~C, Figure~C1).

\begin{table}[t]
\centering
\small
\setlength{\tabcolsep}{4pt}
\caption{Three-condition progression under increasingly reasoning-permitting access ($N=40$ per model-item across all three conditions). Position-lock counts an item as locked when the majority answer letter is identical under both orientations. An exact $A$/$B$ split yields no majority and is \emph{not} counted as locked, since an even split is not evidence of a lock. Two Opus items fall exactly even (one in the raw-API and one in the reasoning-enabled condition), and the figure is sensitive to this choice: breaking those ties toward a letter instead would report $0.44$ and $0.28$ in place of $0.39$ and $0.22$. The analysis script included in the code and data supplement applies the rule stated here.}
\label{tab:progression}
\begin{tabular}{lcccc}
\toprule
model & \makecell{as-dep.\\CLI} & \makecell{raw\\API} & \makecell{reason.\\enabled} & \makecell{position-lock\\(CLI\,$\to$\,raw\,$\to$\,en.)} \\
\midrule
Opus & 0.34 & 0.66 & 0.64 & $0.61\to0.39\to0.22$ \\
Sonnet & 0.48 & 0.79 & 0.67 & $0.50\to0.28\to0.17$ \\
DeepSeek & 0.06 & ---$^\dagger$ & 0.63 & $1.00\to{}$---$^\dagger\to0.22$ \\
\bottomrule
\end{tabular}
\end{table}

\noindent$^\dagger$\emph{DeepSeek-V4-Flash is a reasoning model that returns empty content without a reasoning budget, so a no-reasoning raw reading is undefined; its as-deployed and reasoning-enabled readings bracket the interval.} The reasoning-enabled reading is \textbf{not} asserted as any model's ``true'' concentration; it is a condition under which the instrument measures value-engagement rather than letter-position.

\textbf{Reading Table~\ref{tab:progression} honestly.} The load-bearing pattern is the \textbf{monotone collapse of position-lock} (Opus $0.61\to0.39\to0.22$; Sonnet $0.50\to0.28\to0.17$; DeepSeek $1.00\to0.22$, middle condition unavailable --- Table~1) and the correspondingly large jump of the index off its as-deployed low. Two nuances matter. First, the index \textbf{plateaus} between the raw API and reasoning-enabled conditions (Opus raw $0.66\approx$ thinking 0.64; Sonnet 0.79 vs 0.67): reasoning does not push extremity above its raw-API value. Second, the bulk of the artifact-removal is thus the \textbf{as-deployed $\to$ raw jump}, and reasoning's main \emph{further} effect is to reduce position-lock still further (Opus $0.39\to0.22$, Sonnet $0.28\to0.17$), not to raise extremity. We separate the two movers: the \emph{client} change carries the extremity jump, the \emph{reasoning} change the remaining lock collapse. Both are access-configuration changes, which is our claim; we do not attribute the extremity jump to reasoning suppression, since reasoning is off on both sides of that step. DeepSeek is the sharpest instance --- bare-letter through a constrained-budget client it collapses to a full letter-lock (0.06), while through its raw API with reasoning enabled it content-tracks (0.63); the $0.06\leftrightarrow0.63$ gap is a property of the access path, not the model. Rising off the low reading certifies the low as-deployed reading was not identifiable as a disposition; it does not certify the reasoning-enabled number as canonical.

\textbf{The softness is carried by the locked dilemmas.} The failure localizes item by item, which also shows what these models look like where they \emph{do} engage: splitting each model's 18 dilemmas into position-locked and content-flipping ones, the within-subset extremity separates cleanly (the split for the four models with non-trivial locked subsets is in Appendix~C, Table~C1). The headline ``Opus 0.34, soft flagship'' is an aggregation artifact: on the 11 dilemmas Opus answers by position (within-subset extremity 0.17) it drags the mean down, but on the 7 it engages it is moderately committed (0.62), and Sonnet on its 9 engaging dilemmas is strongly committed (0.87). Neither is soft \emph{where the instrument is actually measuring value}.

\textbf{The direction-flip component is what remains identifiable.} We decompose the apparent per-item pairwise distance into two components. \textbf{Direction-flip (identifiable):} the two models fall on opposite sides of 0.5 with margin --- a genuine value disagreement, \emph{largely robust to the position-lock artifact}, since a position-locked model sits at the boundary and cannot direction-disagree, so any surviving flip involves two content-engaging responses. \textbf{Same-side magnitude residual (not identifiable at the low end):} conflates graded commitment, ambivalence, and position-lock, and we do not read it as a value quantity --- labeling it ``determinism'' is exactly the mis-identification this paper corrects. The \textbf{genuine share} of a pair is the fraction of its raw pairwise distance contributed by direction-flips. Because a flip with one model at 0.55 and another at 0.45 is within sampling noise, we report a \textbf{robust flip} count requiring both margins $|P-0.5|$ to exceed a $\pm 0.1$ dead-band that discards near-boundary (position-lock-suspect) items, doing double duty as a boundary-noise control and the operational insulation of the direction component from the artifact. The result: a non-engaging model's distances are \textbf{pure position artifact} (DeepSeek's zero direction-flips reflect non-engagement, not agreement); committed cross-family pairs \textbf{genuinely disagree and survive the robust test} (Grok-3 carries genuine shares of 73--88\% --- direction robust, magnitude uncertain, wide 18-item CIs --- against GPT-5.5, GPT-5.4, Gemini, Haiku, and Fable 5; at the stricter $\pm 0.16$ band they give 62--88\%, Appendix~C); and the ``mixed middle'' resolves only with the position-lock diagnostic attached. The full per-pair derivation is in Appendix~C.

\textbf{An ordering that survives, reinterpreted.} (Appendix~E: same pattern outside the value set.) One relational fact survives the whole diagnosis, and it is easy to over-read. Across all three access conditions the smaller Haiku is \emph{more} concentrated than the flagship Opus (as-deployed $0.34<0.77$, raw API $0.66<0.85$, reasoning-enabled $0.64<0.76$): ``a small model out-commits the flagship'' is neither a client nor merely a position-lock artifact, holding even where both models content-track. This is \emph{not} evidence that Opus is value-neutral --- it commits moderately where it engages --- only that it concentrates less than Haiku on this item set. \texttt{claude-haiku-4-5} rejects both adaptive thinking and the effort parameter, so it cannot be re-collected under the \emph{same} configuration as the other two. It does support a fixed-budget extended-thinking mode, and we collected that condition ($N=40$ per item, reasoning engaged on 720 of 720 draws), obtaining extremity 0.76 with position-lock falling from 0.111 at the raw API to 0.056 under fixed-budget thinking. Haiku is not a collapse case --- its lock was already near zero as-deployed (0.056), so its trace is 0.056$\to$0.111$\to$0.056, not a monotone decline. The ordering therefore holds on a measured reasoning-enabled reading rather than a carried-forward one, at the cost that the three models are compared each at \emph{its own} maximal reasoning configuration rather than at an identical one. We report the ordering as robust in direction, modest in what it licenses. Within-family individuation is weak and the robust disagreement is cross-family: once the $\pm 0.1$ robust test is applied most apparent same-family Anthropic flips do not survive, and because Opus and Sonnet are themselves substantially position-locked as-deployed, naive same-family flip counts are unreliable in both directions (inflated by boundary noise, deflated by position-lock). The detailed within-family analysis and two under-powered persona axes are in Appendix~D.

\section{5. Access Configuration as One Generator of the Identifiability Failure}

\S4 establishes \emph{that} the index is not identifiable at its low end and that the low readings resolve when reasoning is admitted. This section identifies \textbf{one generator}: the access configuration through which a model is served --- the deployment client and the reasoning-on/off setting it implies. We do \textbf{not} claim the access path ``changes a model's values''; we claim it is one route by which a model is pushed into position-lock, so the \emph{same} model reads value-neutral through one configuration and value-committed through another --- an identifiability hazard for any audit that fixes a single path. In a frozen benchmark this axis is confounded with provider, since each provider was collected through a provider-typical client.

\textbf{The deployment client and the reasoning setting as routes into lock.} Our frozen set was collected through provider-typical clients (Anthropic via the subscription CLI, OpenAI via an OAuth client, Google and xAI via direct API, DeepSeek via a generic agent-runtime client). Re-collecting the same models on the same backend through their raw provider API, we measure how far the profile moves (Table~\ref{tab:access-shift}).

\begin{table}[t]
\centering
\caption{Access-path shift between the frozen-benchmark client and the raw API (fixed model and backend). Opus's 74 refusals (of 720 draws) are dropped per-draw and each item's $P(v_1)$ is recomputed over its remaining valid draws before the shift is computed (not whole-item exclusion).}
\label{tab:access-shift}
\begin{tabular}{llcc}
\toprule
provider & model & \makecell{mean\\$|\Delta P|$} & \makecell{direction\\flips /18} \\
\midrule
Anthropic & Opus & 0.309 & 4 \\
Anthropic & Sonnet & 0.168 & 0 \\
Anthropic & Haiku & 0.224 & 2 \\
Anthropic & Fable 5 & 0.258 & 3 \\
OpenAI & GPT-5.5 & 0.008 & 0 \\
OpenAI & GPT-5.4 & 0.051 & 1 \\
\bottomrule
\end{tabular}
\end{table}

\noindent Anthropic pairs contrast the subscription CLI with the raw API; OpenAI pairs contrast the OAuth client with the raw API. Two facts stand out. \textbf{First, the effect is large for one client:} Opus through the subscription CLI rather than the raw API moves by 0.31 and flips four of eighteen items --- comparable to a genuine cross-model difference --- and all four Anthropic models move (0.17--0.31). For the position-locked models the \emph{content} of that shift is the lock resolving: Sonnet's entire shift is position-lock releasing (its 0.50 lock falls to 0.28 with zero direction flips), Opus's a mixture of lock-release and genuine movement. The access path is not adding a value; it moves the model between a locked (non-identifiable) and an engaged (identifiable) reading. \textbf{Second, the effect is client-specific:} the OpenAI models are near-invariant to \emph{this} client (OAuth-to-raw 0.008 and 0.051), but the \emph{same} GPT-5.5 through a third client (a generic agent runtime) shifts by 0.25 with three flips --- so the statement is not ``OpenAI models are harness-robust'' but ``the OAuth client happens not to induce lock where the subscription CLI does.'' In the frozen benchmark, provider family and access configuration are confounded exactly where the comparison is drawn.

\textbf{The clearest view of the hazard is the concentration index itself.} Re-measuring extremity for the four Anthropic models through the raw API (refusals excluded) gives Opus 0.66, Sonnet 0.79, Haiku 0.85, Fable 5 0.83, against subscription-CLI values 0.34, 0.48, 0.77, 0.79. The most access-distorted number in the paper is the flagship's headline softness (0.34 through the CLI, 0.66 through the raw API) the same model read as far softer through the CLI than through the raw API (neither reading being its true disposition) because the CLI path suppressed its reasoning into partial position-lock. We accordingly treat the \S4 Anthropic as-deployed low extremities as \emph{subscription-CLI-configured} and non-identifiable, not intrinsic. The deeper generator is the \textbf{reasoning-on/off} setting the configuration implies: DeepSeek is starkest --- a reasoning model served bare-letter through a constrained-budget client has its reasoning starved and collapses to a full position-lock, while with reasoning enabled it content-tracks (the $0.06\leftrightarrow0.63$ gap).

\textbf{Reasoning-off is a low-visibility, potentially value-flipping deployment condition} --- where the identifiability failure becomes an alignment hazard rather than a measurement footnote. In real deployments reasoning is frequently \emph{off} for ordinary reasons: a user may not know a model is a reasoning model, or reasoning is disabled to cut token cost, latency, or a serving-compatibility worry. The irony is that a practitioner who disables reasoning \emph{so the model will not behave degenerately} thereby induces the degenerate position-lock. We present this as an illustrative deployment risk with a reproduced mechanism, not a quantified prevalence claim. Reasoning-off is not exotic even \emph{within} a reasoning-enabled configuration: under adaptive thinking at medium and high effort, \texttt{claude-opus-4-8} frequently declined to spend any reasoning tokens on a bare single-letter forced choice, and even at maximum effort it engaged reasoning on well under a quarter of them.\footnote{At maximum effort, reasoning was engaged on 127 of 720 \texttt{claude-opus-4-8} prompts (17.6\%) and 598 of 720 \texttt{claude-sonnet-5} prompts (83.1\%), and it is strongly item-dependent. The two remaining tier/generation cells, the reason we do not read this as a tier or generation effect, and the item-level spread are in Appendix~L.} These models did not reliably reason on a bare-letter forced choice unless pressed to, so the position-lock the frozen instrument recorded is a \emph{faithful} measurement of a very ordinary access condition. Raising the reasoning \emph{effort} level does not move it either: on one later model in English, engagement rose from 0 to 21.8\% while per-item concentration and position-lock both stayed flat (Appendix~J).

\textbf{Prompt language is a further such condition.} We re-collected the Korean condition alongside an English one so that the two would come from the same session, holding the model (\texttt{claude-opus-4-8}) and the reasoning configuration fixed: engagement on the same 18 dilemmas is 150 of 717 draws in Korean and 562 of 720 in English (three timeouts excluded). The Korean figure is an independent re-collection of the condition reported in the footnote above, taken eight days later; the two are statistically indistinguishable (127 of 720 versus 150 of 717, overlapping intervals), which is also the only direct check we have that a closed model served under the same name still behaves the same way. The English prompts are the shorter of the two (median 144 versus 221 input tokens), so the effect runs opposite to a longer-prompt explanation, and holds on 16 of 18 items, none the other way, ruling out item selection. Since translation changes length along with language, we report a prompt-language contrast, not an effect of language alone. Whether the model reasons at all on this probe is therefore itself access-conditioned, which is the claim of this section applied to the probe's own instrumentation.

\textbf{The measurement depends on access-supplied compliance.} Through the raw API, Opus \emph{refuses the forced choice} on 10.3\% of draws (74 of 720), responding that a single letter would distort a situation-dependent judgment or that it is not the actor entitled to decide. The other three Anthropic models refuse 0\%, and the \emph{same} Opus through the subscription CLI refuses 0\%. We do not generalize this to a capability law (Fable 5 is flagship-tier and refused 0\%), but the single-model result is consequential: compliance with the one-letter frame is, for this model, \emph{supplied by the access configuration} rather than fixed. A single-draw CLI pipeline records a crisp choice and never sees the refusal; a parser over raw output can even extract a spurious A/B from a hedged refusal --- the same identifiability failure from the compliance side. We exclude the 74 refusals before recomputing the shifts above.

\textbf{White-box control: the system prompt is causal, and the profile shift replicates.} We ran a white-box control on two models --- raw-API Opus and, as a second-model replication, Sonnet --- each with a representative agent-style system prompt added. Because Opus 4.8 exposes no temperature parameter, the injected system prompt is the only manipulated variable, so the subscription-CLI-to-raw difference cannot be attributed to a researcher-controllable temperature. \textbf{The profile-shift effect replicates at $n=2$:} the system prompt moves Opus by $\Delta P = 0.170$ and Sonnet by $0.126$ --- comparable, so it is not Opus-specific. \textbf{The refusal-elimination effect is Opus-specific:} it eliminates Opus's refusal (0 of 720 vs 85, same-run direct arm; the 10.3\% above is the separate raw-API set), establishing white-box that the access-supplied compliance above is \emph{caused} by the system prompt; Sonnet does not refuse under the bare raw API, so has none to eliminate. We state both parts and imply neither full replication nor a single mechanism. \textbf{Reproduction of the specific client is partial:} the generic prompt does not reproduce the subscription-CLI profile (agent-system vs CLI 0.336, \emph{larger} than raw vs CLI 0.309), moving Opus in \emph{a} direction, not toward that profile. We establish white-box that an agent system prompt causally (a) compels the forced choice (Opus) and (b) shifts the profile (both models), and mark black-box that the \emph{particular} 0.31 subscription-CLI shift is caused by its proprietary, version-bound system prompt.

\textbf{What this means for the low-end reading.} The two sources compose: the identifiability failure (\S4) makes a low concentration reading uninterpretable within a single access path, and the access-configuration result makes the position-lock that produces it \emph{inducible} by an ordinary, often-invisible deployment choice. A cross-model value distance in a frozen benchmark is therefore, at its low end, a distance between readings each conditioned on its access path, and for at least one major provider that path suppressed reasoning into partial lock. The surviving cross-family individuation is real; the naive low-end reading is not. Every figure here is version-bound (Appendix~G). \textbf{Prompt language is a further measurement condition of the same kind} --- re-running on English-translated items leaves the relational ordering intact while shifting the absolute profile (Appendix~F).

\section{6. Discussion}

\paragraph{The alignment stake, sharpened.} The consequence for auditing is sharper than ``a non-committal model evades a distance threshold.'' The models an audit \emph{reads} as non-committal are, in our data, largely not: they are position-locked by the access path. A concentration-blind, single-path audit will thus mislabel a \textbf{lock-suppressed model as value-neutral when it may hold a committed disposition the access configuration rendered invisible} --- a more consequential failure than genuine indifference, since the audit reports absent commitment where the engaged model would commit --- a misclassification risk, not a safety verdict. An auditor running a cost-optimized (reasoning-off) client over a reasoning model reads it as ``value-neutral.''

\paragraph{Every concentration reading needs a position-lock reading beside it.} Because the lock is access-inducible, ``as-served'' must name the deployment client and reasoning setting, and a reading is interpretable only against a named configuration and version (Appendix~G). Reported this way, genuine cross-family disagreement still shows through: Grok against several models, 73--88\%, direction robust and magnitude uncertain.

\paragraph{Access configuration is one generator, not the whole story.} The access path is one route into position-lock, established here for two clients. Other generators (format, item language, Appendix~F) move the profile too. We name it because it is common, low-visibility, and confounded with provider in our frozen benchmark.

\paragraph{From confound to controlled variable.} Fixing and naming the eliciting configuration turns an access-path confound into a controlled intervention, allowing prompts and reasoning settings to be studied against identifiable dispositions. The diagnostic therefore invalidates the undifferentiated low-end reading while preserving the direction-flip signal on engaging items.

\paragraph{The diagnostic, as a procedure.} The result converts into five steps an audit can run. (1)~Measure counterbalanced, both orientations per item. (2)~Per item, record whether the majority answer \emph{letter} is the same under both orientations. (3)~Report the resulting position-lock fraction \emph{beside} every concentration reading, never instead of it. (4)~Treat a low concentration value as uninterpretable until position-lock is excluded to a pre-specified tolerance: compare the lock fraction's uncertainty interval against the chance floor (Appendix~C), since failing to detect a lock is not evidence of its absence. (5)~Where the lock is non-trivial, report the subset-conditional concentration among the engaging items --- informative about where the instrument did measure, but a conditional quantity, not an unbiased substitute for the aggregate.

\section{7. Limitations}

\textbf{The access-configuration result is one generator, not a path-matched census.} A fully path-matched, reasoning-enabled collection across all nine models (the correct baseline for future revision) is not yet done; the OpenAI models are clean \emph{through the OAuth client specifically} (the same GPT-5.5 shifts $\sim$0.25 through a third client). The white-box control shows a system prompt is sufficient to shift the profile (two models) and eliminate the refusal (Opus) but does not reconstruct the proprietary subscription-CLI prompt, so the \emph{particular} deployed shift stays black-box.

\textbf{We do not claim a ``true'' value, or a prevalence.} The reasoning-enabled reading is not a ground-truth disposition; there is no single ground truth, and readings are measurement-conditioned. Our assertion is asymmetric: position-lock is an \emph{invalid} reading (measuring letter-position), while content-engaging readings are condition-dependent but valid. The alignment-hazard argument correspondingly rests on a reproduced mechanism plus an illustrative deployment risk; we make no numerical claim about how common reasoning-off is.

\textbf{The same-side magnitude residual is not identified, and the per-model ordering is under-sampled} (Appendix~I).

\section{8. Conclusion}

Cross-model value comparisons rest on a counterbalanced forced-choice instrument whose concentration index is \emph{not identifiable at its low end}: counterbalancing maps a position-lock --- a content-independent answer-letter bias --- onto the same near-0.5 signature as genuine neutrality. Across nine models the position-lock fraction tracks the index at $r = -0.986$ (descriptive, structurally constrained, \S4); the three models read as soft are the position-locked ones, and their softness resolves as the access path admits reasoning --- an identifiability failure, not a disposition. We claim not a ``true'' value but that the index alone cannot identify commitment there, so a position-lock diagnostic must accompany every reading. Access configuration is one generator, confounded with provider in a frozen benchmark. What remains identifiable is the direction-flip component.

\section*{Data and Code Availability}

All response-level data and every analysis script used to produce the numbers in this paper are
supplied as a self-contained code-and-data package, attached to this arXiv submission as
\textbf{ancillary files}. The package contains the raw and parsed responses for the frozen nine-model set and for
each re-collection condition, the analysis scripts named throughout these appendices, a
\texttt{CHECKSUMS} manifest over every file, and a single \texttt{verify.sh} that re-runs the
result reproductions end to end without network access or API keys. Where an appendix refers to a script or file ``included in the code and data
supplement,'' it refers to that package.

Two limits are stated plainly. First, the pinned client versions and build identifiers are given in
Appendix~G, and every figure is to be read against them rather than as a client-independent
constant. Second, re-running the
\emph{collection} rather than the analysis requires provider API keys and will not reproduce the
frozen rows draw-for-draw; what must replicate is the structure, not the individual draws.

\section*{Appendix A. Extended Related Work}

\textbf{Counterbalancing is a debiasing tool that creates the failure, not a competitor to it.} Closest to our diagnosis, \citet{dominguez2023survey} show that when survey-style option order is randomized, LLMs' apparent response distributions collapse toward balanced, exposing position bias as a large component of naive cross-model differences. Order-randomization is the standard fix, and our counterbalanced measurement \emph{is} such a correction. Our point is that this fix, once its output is summarized by a concentration index, introduces a second-order identifiability problem the position-bias literature does not address: a position-lock --- a \emph{residual} position bias that survives label-swapping because the model tracks the letter, not the option --- is mapped by the very act of counterbalancing onto the same near-0.5 signature as genuine neutrality. The debiasing tool hides this specific bias rather than exposing it.

\textbf{Concentration/decidedness has been measured, but only within a model, and read as a trait.} A separate line quantifies how sharply or consistently a \emph{single} model commits: the decidedness or ambiguity of a model's moral choices (MoralChoice's decidedness split~\citep{scherrer2023moralchoice}), repeated-sampling agreement and answer-consistency (TARr@N/TARa@N~\citep{atil2024nondeterminism}), value consistency across paraphrase, use-case, and model (Moore et al.~\citep{moore2024consistent}), and the persistence of a value leaning under persona prompting~\citep{lee2024inertia}. A forced-choice value-profiling study across three providers (Hadar-Shoval et al.~\citep{hadarshoval2024alignment}) is the closest in setup, reporting per-model value profiles; we cite it for that cross-provider profiling and not as evidence of within-model commitment, since its own reliability analysis reports run-to-run \emph{variation} (ICC $\approx 0.85$ over ten runs). Closest to our aim, \citet{kriegmair2026individuality} explicitly separate \emph{genuine} model individuality from a response-style confound across ten models, using crossed random-effects (variance-components) models on 74.9M psycholinguistic norm ratings to partition, per model, stimulus-specific individuality from global response bias and noise. We share their diagnosis that a response-style confound inflates apparent model individuality, but the instrument differs: they partition \emph{within-model} variance, whereas we identify a \emph{non-identifiability} of a counterbalanced concentration index and decompose a \emph{pairwise between-model distance}; their data are continuous norm ratings, ours counterbalanced repeated forced choices; and their ``response bias'' is a global rating tendency, whereas our artifact is a content-independent letter-lock that counterbalancing averages into apparent neutrality. What none of this work does is recognize that the counterbalanced concentration index cannot, at its low end, tell a committed-but-lock-suppressed model from a genuinely balanced one.

\textbf{Adjacent: persona and identity drift.} A related literature frames within-context deviation of a \emph{stated persona} as a defect and targets its prevention. \citet{li2024instability} benchmark persona degradation over multi-round dialog and propose a decoding intervention to restore stability; \citet{kim2024identitydrift} find across nine models that assigning a persona does not prevent identity drift. Our object is different: the \emph{base} value disposition of a model given no persona, and the identifiability of the instrument that reads it. Similarly, work on sycophancy and steerability \citep{sharma2023sycophancy, denison2024subterfuge} bears on the auxiliary persona-override axis we defer to Appendix~D, not on the main result.

\textbf{The access axis: elicitation and harness sensitivity.} Two literatures bear on \S5. One shows that \emph{how} a model is invoked (its scaffold, prompt formatting, and CLI harness) materially changes measured behavior, and argues agent comparisons are uninterpretable without disclosing the harness (Sclar et al.\ on format sensitivity~\citep{sclar2024formatspread}; ``Stop Comparing Language-Model Agents Without Disclosing the Harness''~\citep{zhang2026harness}; Harness-Bench~\citep{yao2026harnessbench}). That work concerns \emph{task and agent performance}, not value elicitation, and does not connect a harness to an identifiability failure of a value estimator. The other strand is nearer: \citet{rottger2024political} show that models give substantively different value answers when not forced into a multiple-choice format, that the answer depends on \emph{how} the choice is forced, and that responses lack paraphrase robustness --- a direct precedent for our claim that the elicitation frame shapes what is measured, and for the forced-choice refusal of \S5. ``Mind the Gap''~\citep{mahajan2026mindthegap} likewise varies the elicitation protocol and studies how it moves a stated-versus-revealed preference gap. A forced-choice study~\citep{chen2026choices} finds that reformulating a prompt as a constrained multiple choice \emph{induces} compliance and policy-violating selections that open-ended prompts do not; compliance-as-induced is thus already on record, in a safety-violation construct. Our contribution is the intersection none of these occupy (see \S2). Provider-layer steering --- system-prompt sycophancy and steerability~\citep{sharma2023sycophancy, denison2024subterfuge} --- concerns the model owner's prompt, not the third-party access path, and is adjacent rather than the opponent.

\textbf{Construct grounding.} Dilemma axes are grounded in Schwartz's theory of basic human values and in Moral Foundations Theory, operationalized through forced-choice value inventories in the tradition of the PVQ-RR. The item-to-axis mapping is in Appendix~H.

\section*{Appendix B. Provenance and Data Integrity}

Each response stores the model's \emph{raw} output alongside the parsed choice. This raw store is not cosmetic. During collection it was the sole basis for detecting three data-quality incidents before the dataset was frozen: two binary-swap failures from an auto-update mid-run, and a parsing contamination on one provider path that produced a \emph{spurious} soft reading for one model (a false extremity of 0.11 for Gemini) which, on re-collection through a clean path, resolved to its near-fully-concentrated value (0.80). This parser-garbage incident is a \emph{different} phenomenon from the position-lock the paper is about: parser contamination is a scoring failure that fabricates a soft reading from a normal response, whereas position-lock is a genuine, reproducible model \emph{behavior} (a content-independent letter choice) that counterbalancing then averages into apparent softness. Had only parsed choices been stored, the parser incident would have been invisible; the position-lock, by contrast, is visible in the raw letters and is precisely what the \S3 diagnostic measures. We report the incident log as part of the datasheet, as evidence of the dataset's integrity. (A late-stage data-integrity note: the $N=40$ reasoning-enabled re-collection was run after a runner fix --- \texttt{max\_tokens = 16000}, a letter-\texttt{None} guard on error rows, and exclusion of the no-thinking-mode Haiku --- with the earlier contaminated partial isolated and excluded.)

\section*{Appendix C. Per-Model Item Split and the Direction-Flip Derivation}

\begin{table}[h]
\centering
\caption*{\textbf{Table C1.} Per-model split of the 18 dilemmas into position-locked vs.\ content-engaging items, with within-subset extremity (as-deployed).}
\setlength{\tabcolsep}{4pt}
\begin{tabular}{lcc}
\toprule
model & \makecell{position-locked\\dilemmas (extremity)} & \makecell{content-engaging\\dilemmas (extremity)} \\
\midrule
Opus & 11 (0.17) & 7 (0.62) \\
Sonnet & 9 (0.09) & 9 (0.87) \\
Haiku & 1 (0.05) & 17 (0.82) \\
GPT-5.5 & 0 (---) & 18 (0.947) \\
\bottomrule
\end{tabular}
\end{table}

\begin{table}[h]
\centering
\caption*{\textbf{Table C2.} Extremity and position-lock, nine models, as-deployed (frozen set, $N=40$ per model-item; the data plotted in Figure~2 of the main paper). Position-lock fraction is the fraction of the 18 dilemmas answered with the same letter under both orientations. The first two columns are near-collinear ($r=-0.986$, descriptive; the two quantities are structurally constrained, \S4 of the main paper).}
\small\setlength{\tabcolsep}{3pt}
\begin{tabular}{lccp{2.2cm}}
\toprule
model & extremity & \makecell{position-lock\\fraction} & reading \\
\midrule
DeepSeek-V4-Flash & 0.06 & 18/18 & full lock (non-engaging) \\
Opus & 0.34 & 11/18 & partial lock (``soft'') \\
Sonnet & 0.48 & 9/18 & partial lock \\
Haiku & 0.77 & 1/18 & content-tracking \\
GPT-5.4 & 0.78 & 4/18 & partial tracking \\
Grok-3 & 0.79 & 2/18 & content-tracking \\
Fable 5 & 0.79 & 3/18 & content-tracking \\
Gemini-2.5-flash & 0.80 & 1/18 & content-tracking \\
GPT-5.5 & 0.95 & 0/18 & fully tracking \\
\bottomrule
\end{tabular}
\end{table}

The headline ``Opus 0.34, soft flagship'' is an artifact of aggregation. On the 11 of 18 dilemmas Opus answers by position (extremity 0.17) it drags the mean down, but on the 7 it actually engages it is moderately committed (0.62); Sonnet on its 9 engaging dilemmas is strongly committed (0.87). GPT-5.5 has \emph{no} position-locked items --- all 18 dilemmas engage --- so its engaging-subset extremity equals its overall 0.947, and its locked-subset cell is undefined (---). Haiku's single locked item is worth naming, because \emph{which} item it is depends on the parser and that dependence is this paper's subject. Under the collector's loose parse the locked item was E02, where 26 of Haiku's 40 draws are refusals or hedges (``I cannot reduce this choice to A or B'') from which the fallback extracted a letter; under the strict rule shipped with the code supplement (\texttt{analysis/parse\_rule.py}) those draws are not choices, E02 ceases to look locked, and the locked item is D-new-03 at extremity 0.05. The loose parse did not merely add noise --- it manufactured a position-lock out of a model declining to answer, which is the failure mode the main paper attributes to the instrument, occurring inside our own instrument. Neither Anthropic model is soft \emph{where the instrument is actually measuring value}. This is the concrete form of the identifiability failure: the aggregate index reports a disposition that the item-level split shows to be a mixture of a moderate-to-strong genuine commitment and a content-independent lock.

\textbf{A non-engaging model's distances are pure position artifact.} DeepSeek-V4-Flash presents as extremity 0.06, but the letter-invariance check shows this is not a balanced disposition: the model selects the ``B'' label on 95\% of draws and its letter does \emph{not} flip when labels swap (letter-invariant on all 18 items, versus GPT-5.5's content-tracking flip on all 18), a non-engagement of the kind reported for smaller models~\citep{pinhanez2025nondeterminism}. Its zero direction-flips against every other model (eight pairwise distances, range 0.16--0.47, genuine share 0\%) reflect non-engagement, not value agreement: a single draw would record a definite ``B'' and mistake it for a committed choice. This is the extreme case of the identifiability failure.

\textbf{Committed model pairs genuinely disagree, and it survives the robust test.} At the other end, Grok-3 carries real directional differences from most models, with genuine shares of 73--88\% against GPT-5.5, GPT-5.4, Gemini, Haiku, and Fable 5. These are point estimates over 18 items with wide 95\% bootstrap intervals (e.g., Grok/Gemini 88\% [70, 98]; Grok/Fable 5 73\% [38, 92]), so they indicate a robust \emph{direction} of disagreement rather than a precise magnitude. Under the $\pm 0.1$ dead-band the leading pairs hold (GPT-5.5/Grok retains ten robust flips, Haiku/Grok eight), and Fable 5 disagrees in direction with the OpenAI models and Gemini at high genuine shares (Fable 5/GPT-5.5 86\%, Fable 5/Gemini 87\%, the latter holding nine robust flips). Genuine value individuation between content-engaging models is real and concentrated \emph{across} provider families rather than within them (Appendix~D).

\textbf{The mixed middle.} GPT-5.5 against the Anthropic mean divides into 5 direction-flips (genuine), 8 same-side items on which GPT-5.5 is more extreme by more than 0.2, and 5 remaining items (2 with no Anthropic majority direction, 3 same-side with a smaller magnitude gap), summing to 18, and the same ``GPT-5.5 vs Anthropic'' divergence is mostly genuine against one sibling and mostly same-side against another (Opus/GPT-5.5 38\% genuine, Haiku/GPT-5.5 72\% genuine), because the two siblings sit at very different extremities. As \S4 shows, Opus's low extremity is itself substantially position-lock. Only the decomposition, with the position-lock diagnostic attached, exposes this; a family-averaged raw distance hides it.

\begin{figure}[h]
\centering
\includegraphics[width=0.85\linewidth]{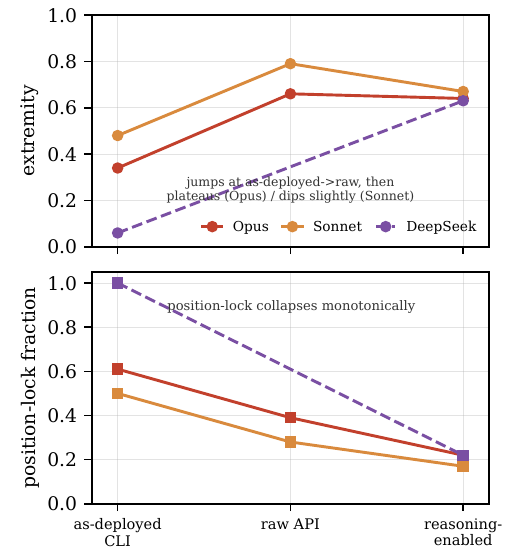}
\caption*{\textbf{Figure C1.} Three-condition progression (as-deployed CLI $\to$ raw API $\to$ reasoning-enabled), plotting Table~1 of the main paper. Position-lock (bottom) collapses monotonically; extremity (top) jumps at the as-deployed$\to$raw step, then plateaus (Opus $0.66\approx0.64$) or dips slightly (Sonnet $0.79\to0.67$) while lock keeps collapsing. DeepSeek's raw-API point is undefined (dashed bridge).}
\label{fig:prog-supp}
\end{figure}

\textbf{The chance floor of the position-lock diagnostic.} The diagnostic asks whether a model's
majority answer letter is the same under both orientations. A genuinely indifferent,
content-tracking model also answers 50/50, so it reproduces its majority letter by chance. With
20 draws per orientation, $P(\text{majority A}) = P(\text{majority B}) = 0.412$ and
$P(\text{tie}) = 0.176$, so $P(\text{same non-tie majority}) = 2(0.412)^2 = 0.339$: about
\textbf{6 of 18 items} are expected to read as locked with no lock present. Against that floor
(binomial, $n=18$, $p=0.339$, $\mathrm{sd} = 2.0$), DeepSeek's 18/18 gives $z = +5.9$ and Opus's
11/18 gives $z = +2.4$; \textbf{Sonnet's 9/18 gives $z = +1.4$ and is not separable from
indifference.} We therefore treat DeepSeek and Opus as established position-lock and Sonnet as
undetermined, and we note that the collapse figures should be read against this floor rather than
against zero. Raising $N$ or replacing the majority vote with an exact per-item test on the
letter split would sharpen this; both are analysis, not collection.

\textbf{The parse rule, before and after, with the denominators.} Every number in this paper
resolves its forced choices through one rule (\texttt{analysis/parse\_rule.py} in the code
supplement): a response counts as a choice only if it \emph{begins} with \texttt{A} or
\texttt{B}. Until 2026-07-28 the main analysis instead read the collector's stored parse, whose
fallback took any \texttt{A} or \texttt{B} appearing anywhere in the response --- including inside
prose and inside API error strings. Table~C3 reports both, because the difference is small enough
to state completely and consequential enough that a reader should not have to take our word for it.

\begin{table*}[t]
\centering
\small
\setlength{\tabcolsep}{4pt}
\caption*{\textbf{Table C3.} Effect of the parse rule on the as-deployed set, with valid
denominators. ``Dropped'' are draws whose raw text is not a forced choice under the rule; they are
dropped, never imputed. Grok's raw output is withheld under its provider's terms (Appendix~B), so
its rows keep their stored value and cannot move. The two extremity columns are before and after the parse rule, not a progression: an arrow in this paper means an access-path condition changed, and nothing about the access path changed here. Position-lock is identical under both parses, so it gets one column.}
\begin{tabular}{lcccccc}
\toprule
model & collected & used & dropped & \makecell{extremity\\stored} & \makecell{extremity\\rule} & \makecell{lock\\(both)} \\
\midrule
Opus              & 720 & 719 & 1  & 0.344 & 0.343 & 0.611 \\
Sonnet            & 720 & 720 & 0  & 0.483 & 0.483 & 0.500 \\
Haiku             & 719 & 679 & 40 & 0.732 & \textbf{0.774} & 0.056 \\
Fable 5           & 720 & 720 & 0  & 0.789 & 0.789 & 0.167 \\
GPT-5.5           & 720 & 720 & 0  & 0.947 & 0.947 & 0.000 \\
GPT-5.4           & 720 & 720 & 0  & 0.778 & 0.778 & 0.222 \\
Gemini-2.5-flash  & 720 & 720 & 0  & 0.800 & 0.800 & 0.056 \\
DeepSeek-V4-Flash & 720 & 720 & 0  & 0.064 & 0.064 & 1.000 \\
Grok-3 (withheld) & 720 & 720 & 0  & 0.794 & 0.794 & 0.111 \\
\midrule
\multicolumn{4}{l}{correlation $r$} & $-0.980$ & $\mathbf{-0.986}$ & --- \\
\bottomrule
\end{tabular}
\end{table*}

One model moves and the rest do not. All 40 dropped draws are Haiku's, and 26 of them fall on a
single item (E02, itself collected at 39 draws rather than 40), where the model declined the format
outright: ``I cannot reduce this choice to A or B'' and ``I appreciate the clear question, but I
should be honest: I'm an AI, not a person.'' The loose fallback extracted a letter from that prose
and counted it as a choice.

\textbf{This changes which item Haiku's lock is.} Under the stored parse Haiku's one locked item is
E02, at within-item extremity 0.23; under the rule E02 has only 13 valid draws, ceases to read as
locked, and the locked item becomes D-new-03 at 0.05 --- the figures in Table~C1. We record this
plainly because it is the paper's own thesis turned on the paper: a parser that manufactures
choices out of refusals manufactured a position-lock that was not there. The instrument being
audited here included our own.

\textbf{Robustness of the correlation to model selection.} The nine models are a convenience
selection, four share one provider, and there is no sampling frame of models to generalize to, so
the Fisher-z interval in \S4 of the main paper is a sensitivity interval rather than a population
one. The relevant question is whether the relationship depends on any particular model or family.
It does not.

\begin{table}[h]
\centering
\small
\caption*{\textbf{Table C4.} Correlation between extremity and position-lock under model exclusions.
``One per provider'' takes a single Anthropic and a single OpenAI model, with Gemini, DeepSeek and
Grok, over all eight such combinations.}
\begin{tabular}{lcc}
\toprule
subset & $n$ & $r$ \\
\midrule
all models                        & 9 & $-0.986$ \\
leave-one-model-out (range)       & 8 & $-0.966$ to $-0.991$ \\
\quad worst case: without DeepSeek& 8 & $-0.966$ \\
leave-one-family-out (range)      & 5--8 & $-0.966$ to $-0.993$ \\
\quad without Anthropic (4 models)& 5 & $-0.991$ \\
one per provider (8 combinations) & 5 & $-0.987$ to $-0.995$ \\
\bottomrule
\end{tabular}
\end{table}

Dropping DeepSeek --- the single most extreme point, and the one a reader is most likely to suspect
of driving the fit --- leaves $r = -0.966$. Removing the entire Anthropic family, which is the
non-independence a reviewer should worry about most, leaves $-0.991$. Every subset that keeps one
model per provider, the most independent reading available in this data, lies between $-0.987$ and
$-0.995$. The relationship is a property of the estimator, not of the sample; as \S4 argues, that
is expected, which is why we read the correlation descriptively and put the paper's weight on the
re-measurement result instead.

\textbf{Dead-band sensitivity.} The robust-flip test of \S4 uses a $\pm 0.1$ dead-band, which is
deliberately looser than the 95\% half-width at $P=0.5$ ($\approx 0.16$ at $N=40$). At $\pm 0.16$
the Grok-3 genuine shares quoted in \S4 fall from 73--88\% to 62--88\%, and the robust-flip count
over all 36 model pairs falls from 136 to 113. The direction of disagreement survives the stricter
screen throughout; only the floor moves.

\section*{Appendix D. Within-Family Individuation and Auxiliary Persona Axes}

\textbf{Within-family individuation is weak, and was under-claimed.} Same-provider comparisons show far less genuine individuation than a naive flip count suggests, once the robust test is applied. Within the Anthropic family most apparent same-family flips do not survive it: Opus/Sonnet, with seven naive direction-flips, retains only two under the $\pm 0.1$ dead-band, and Opus/Haiku falls from eight to five. Because Opus and Sonnet are themselves substantially position-locked as-deployed (\S4), a chunk of their apparent same-family flips sit at the boundary \emph{because one side is locked}, not because the two models genuinely agree, so the as-deployed within-family picture, if anything, \emph{under-states} the genuine individuation between the content-engaging portions of these models. Re-measured with reasoning enabled, where their locks collapse, more of their engaging dilemmas become available to the direction test. We therefore state the within-family conclusion conservatively: the \emph{robust, de-locked} within-family individuation in this dataset is small, the robust value differences are cross-family, and the naive same-family flip count is unreliable in both directions, inflated by boundary noise and deflated by position-lock.

\textbf{Two auxiliary axes, kept out of the main claims.} The protocol can also measure whether a declared value-persona overrides the base disposition, and whether an installed persona resists counter-pressure. We probed both only on the Anthropic family at $N=10$ per condition. That sample is too small to carry weight, and this paper's claims concern the base measurement's identifiability. We describe the instruments and preliminary observations here and leave properly-powered study to future work: a value-persona-override probe against the neutral-persona control of \S3, and a persona-resistance probe under counter-pressure; both at $N=10$ on the Anthropic family, described as instruments and preliminary observations rather than results.

\section*{Appendix E. An Illustrative Second-Domain Probe}

To ask whether the Opus-below-Haiku ordering is peculiar to the value domain, we re-ran the two models on a second, structurally different set of eight forced binary choices with no right answer (aesthetic-taste preferences: a caf\'e name, a color pairing, a melody). The ordering was preserved (Opus below Haiku), though absolute levels fell (Opus $0.34\to0.03$, Haiku $0.77\to0.53$). This is a single illustrative probe, \emph{not} evidence of domain-generality: one model pair on one additional domain, and Opus's 0.03 aesthetic extremity cannot be distinguished from calibrated indifference --- nor, without a position-lock reading for this domain, from a position-lock; we report the ordering without interpreting its generator. A genuine cross-domain claim would require many models across several domains.

\section*{Appendix F. Prompt Language as a Further Measurement Condition}

Access configuration is not the only condition outside the weights that moves the profile. Prompt \emph{language} is another. We re-ran the full protocol with English-translated items (all 18 dilemmas, $N=40$, each model on its client-matched raw or direct path so language is the only change) for the eight engaging models. Table~F1 reports the per-model comparison.

\begin{table*}[t]
\centering
\caption*{\textbf{Table F1.} Korean$\leftrightarrow$English comparison, each model on its client-matched path.}
\begin{tabular}{lcccccc}
\toprule
model & \makecell{KO\\extr.} & \makecell{EN\\extr.} & \makecell{$\Delta$\\extr.} & \makecell{KO$\leftrightarrow$EN\\mean $|\Delta P|$} & \makecell{dir.\\flips/18} & \makecell{EN\\refusal \%} \\
\midrule
Opus & 0.66 & 0.82 & $+0.15$ & 0.20 & 2 & 5.1 \\
Sonnet & 0.79 & 0.81 & $+0.02$ & 0.15 & 4 & 0.0 \\
Haiku & 0.85 & 0.98 & $+0.13$ & 0.20 & 3 & 0.0 \\
Fable 5 & 0.83 & 0.86 & $+0.03$ & 0.21 & 5 & 0.6 \\
GPT-5.5 & 0.96 & 0.87 & $-0.09$ & 0.22 & 5 & 0.0 \\
GPT-5.4 & 0.77 & 0.93 & $+0.16$ & 0.26 & 6 & 0.0 \\
Gemini & 0.80 & 0.78 & $-0.02$ & 0.19 & 4 & 0.0 \\
Grok & 0.79 & 0.79 & $-0.00$ & 0.15 & 3 & 0.0 \\
\bottomrule
\end{tabular}
\end{table*}

\noindent\emph{(KO extremity is the client-matched raw/direct baseline, not the as-deployed value of \S4; for GPT-5.5 it reads 0.96 raw here versus 0.95 through the OAuth client in Table~C2, the small OAuth-to-raw gap of \S5. $\Delta$ extremity is computed from unrounded values. DeepSeek is omitted as position-locked.)}

The specific Opus--Haiku ordering, and Opus's position as the lowest-concentration Anthropic model, are preserved across the two prompt languages. The ordering that carries \S4 is preserved: Opus remains the least-concentrated Anthropic model and is out-committed by Haiku (English 0.82 vs 0.98), and the flagship's refusal recurs in English (Opus 5.1\%; the Korean figure of 10.3\% comes from the earlier 2026-07-11 raw-API session, so this is a cross-session comparison and we do not read the difference in rate), so it is not a Korean artifact. The \emph{absolute} profile is not language-invariant: the mean profile moves by $|\Delta P| \approx 0.20$ between languages (range 0.15--0.26, two to six flips), and extremity is generally higher in English. Language is thus a measurement condition of the same kind as access configuration --- it reshapes the elicited profile while leaving the relational structure intact. What is comparable across deployments is the \emph{decomposed, relational, position-lock-diagnosed} structure, not the absolute low-end reading, which every condition reshapes.

\section*{Appendix G. Toolchain and Version Pinning}

The access effect of \S5 is a function of the client \emph{version}; the concentration and decomposition results of \S3--\S4 are likewise measured as-configured. All figures hold at the following pinned toolchain and are to be re-measured when a client updates.

\begin{itemize}
\sloppy
\item \textbf{Anthropic subscription CLI} (\texttt{claude -p}): Claude Code 2.1.206 (measured 2026-07-11).
\item \textbf{OpenAI codex client:} openclaw 2026.4.27 (build cbc2ba0).
\item \textbf{Agent-runtime client} (access-path cross-checks; DeepSeek frozen path): hermes v0.18.2.
\item \textbf{Raw provider APIs:} Anthropic \texttt{anthropic-version:\allowbreak{} 2023-06-01}; OpenAI Chat Completions; xAI and Google direct API.
\item \textbf{Reasoning-enabled re-collection (\S4):} Anthropic Messages API with adaptive extended thinking at maximum effort, $N=40$ per model-item over all 18 dilemmas; Opus and Sonnet only (\texttt{claude-haiku-4-5} rejects both adaptive thinking and the effort parameter, so this configuration cannot be applied to it; its fixed-budget extended-thinking mode \emph{was} collected separately (\texttt{budget\_tokens = 1024}, $N=40$ per item, reasoning engaged on 720 of 720 draws) and is reported in \S4, so the comparison across models is one of each model's maximal available reasoning condition rather than one identical setting). DeepSeek reasoning-enabled via its raw API with reasoning permitted (\texttt{max\_tokens = 4096}).
\item \textbf{Sonnet white-box control (\S5):} raw-API Sonnet with and without the representative agent-style system prompt.
\item \textbf{Model snapshots:} Anthropic \texttt{claude-opus-4-8}, \texttt{claude-sonnet-5}, \texttt{claude-haiku-4-5}, \texttt{claude-fable-5}; OpenAI \texttt{gpt-5.5}, \texttt{gpt-5.4}; Google \texttt{gemini-2.5-flash}; DeepSeek \texttt{deepseek-v4-flash}; xAI \texttt{grok-3}. The supplementary probes reported in \S5 and Appendices~J, K, and~L additionally use \texttt{claude-opus-5} and \texttt{claude-sonnet-4-6}; neither contributes any row to the nine-model set.
\item \textbf{Sampling.} Opus 4.8 exposes no temperature parameter (deprecated), so the subscription-CLI-to-raw sampling difference is not a researcher-controllable variable; other clients use temperature as-configured (reported, not tuned). Raw-API cross-checks use \texttt{max\_tokens = 2048} (needed for reasoning-style models); the reasoning-enabled re-collection uses the provider's adaptive-thinking budget (\texttt{max\_tokens = 16000}); the reasoning-effort sweep of Appendix~J uses \texttt{max\_tokens = 8192}; and the tier-by-generation batch of Appendices~K and~L uses \texttt{max\_tokens = 8000}. These budgets are pinned here because they are themselves a measurement condition: Appendix~K declines to merge its cells with the \S4 progression precisely because 8000 and 16000 are different conditions. Prompt template held fixed: forced single-letter output, no explanation, except where reasoning is explicitly permitted.
\item \textbf{Collection dates:} frozen nine-model set 2026-07-08--09; access-configuration and raw-API re-collection 2026-07-11; reasoning-enabled and Sonnet-control re-collection 2026-07-17; reasoning-effort sweep, tier-by-generation engagement cells, prompt-language contrast, and the Haiku fixed-budget condition 2026-07-25. In the Korean arm of the language contrast, three draws on a single dilemma failed with network timeouts and were excluded rather than imputed, giving 717 rather than 720 draws.
\end{itemize}

Every \S5 magnitude is to be read as ``$\Delta$ @ \{client version, model snapshot, date\}'' rather than as a client-independent constant.

\section*{Appendix H. Construct Grounding}

The 18 items were authored by us. Item identifiers carry the trace of how the set was built rather than a single consecutive sequence. The core eight are \texttt{D03}, \texttt{D04}, \texttt{D05} and \texttt{D-new-01}, \texttt{-03}, \texttt{-07}, \texttt{-09}, \texttt{-12} --- survivors of a larger authored pool, so their numbers are not contiguous --- and the extended ten are \texttt{E01}--\texttt{E10}. We keep the identifiers as collected so that every identifier in the supplied response files resolves to the same item it did at collection time. Each poses a binary trade-off between two named value poles;
$v_1$ is the pole the analysis scores as the ``first'' option and $v_2$ its counterpart. The poles
were chosen with reference to Schwartz's basic human values and to Moral Foundations Theory, in
the sense that we used those vocabularies to select trade-offs that are recognizable and mutually
exclusive. \textbf{We did not administer a validated instrument, and these items are not a
psychometric scale}: no PVQ-RR item was used verbatim, no factor structure was estimated, and no
inter-rater agreement on the axis labels was collected. The mapping below is therefore a statement
of authorial intent about what each item contrasts, not a validated construct assignment, and the
paper's claims are about the \emph{estimator} applied to these items rather than about the value
constructs themselves.

\begin{center}
\footnotesize
\setlength{\tabcolsep}{3pt}
\begin{tabular}{llll}
\toprule
item & $v_1$ & $v_2$ & set \\
\midrule
\texttt{D03} & consistency & mercy & core-8 \\
\texttt{D04} & fairness & aggregate & core-8 \\
\texttt{D05} & confidentiality & care & core-8 \\
\texttt{D-new-01} & loyalty & fairness & core-8 \\
\texttt{D-new-03} & candor & care & core-8 \\
\texttt{D-new-07} & near-term & long-term & core-8 \\
\texttt{D-new-09} & self-direction & humility & core-8 \\
\texttt{D-new-12} & stimulation & security & core-8 \\
\texttt{E01} & power/status & egalitarian & extended-10 \\
\texttt{E02} & hedonism & discipline & extended-10 \\
\texttt{E03} & conformity & nonconformity & extended-10 \\
\texttt{E04} & tradition & innovation & extended-10 \\
\texttt{E05} & achievement & affiliation & extended-10 \\
\texttt{E06} & prudence & boldness & extended-10 \\
\texttt{E07} & autonomy & interdependence & extended-10 \\
\texttt{E08} & honesty-bluntness & diplomacy & extended-10 \\
\texttt{E09} & competition & cooperation & extended-10 \\
\texttt{E10} & order & flexibility & extended-10 \\
\bottomrule
\end{tabular}
\end{center}

Both sets are included in full in the code and data supplement (see Data and Code Availability); the supplied files carry the same \texttt{axis},
\texttt{v1}, and \texttt{v2} fields used here, so the mapping is checkable against the data rather
than only asserted.

\section*{Appendix I. Extended Limitations}

This appendix expands the compact limitations of \S7 of the main paper. \textbf{The same-side magnitude residual is not identified, and the per-model ordering is under-sampled.} The direction-flip component is largely robust to position-lock and carries our load-bearing claims; the same-side residual conflates graded commitment, ambivalence, and lock. The Opus-below-Haiku ordering comes from a single densely-sampled family, so we can show it survives de-locking but not whether concentration is a per-model trait or a within-provider pattern.

\textbf{Path-matching detail.} Client-matching in \S5 covers the Anthropic (subscription CLI vs.\ raw API) and OpenAI (OAuth client vs.\ raw API) pairs; the OpenAI models are clean \emph{through the OAuth client specifically}, not client-robust in general (the same GPT-5.5 shifts $\sim$0.25 through a third, generic-agent-runtime client). A fully path-matched, reasoning-enabled collection across all nine models is not yet done and is the correct baseline for a future revision.

\textbf{Sampling of the direct-API models.} Gemini and Grok (direct API) were run at default sampling; a temperature spot-check on them left extremity essentially unchanged ($0.80\to0.80$, $0.79\to0.79$). DeepSeek was served through a harness client.

\textbf{Reasoning re-collection matching.} The reasoning-enabled re-collection is $N=40$ (matched to the frozen set) for Opus and Sonnet; the extremity index plateaus between the raw API and reasoning-enabled conditions, so we read the reasoning-enabled column for the further collapse of position-lock rather than for a higher extremity. Haiku cannot be re-collected under the configuration used for the other two, because \texttt{claude-haiku-4-5} rejects both adaptive thinking and the effort parameter. It does support a fixed-budget extended-thinking mode, which we collected ($N=40$ per item, 18 dilemmas, reasoning engaged on 720 of 720 draws): extremity $0.76$, position-lock $0.111\to0.056$, where $0.111$ is its raw-API value. Haiku is not a collapse case and we do not present it as one: its lock was already near zero as-deployed ($0.056$), so its trace across the three conditions is $0.056\to0.111\to0.056$ rather than the monotone decline of Opus and Sonnet, and its extremity moves $0.77\to0.85\to0.76$ rather than plateauing. What the fixed-budget condition establishes for Haiku is only that its high concentration is not an artifact of reasoning being suppressed. The comparison across the three models is therefore each model at \emph{its own} maximal reasoning configuration, not at an identical one; we flag that asymmetry rather than treating the three columns as matched. All \S5 figures are version-bound (Appendix~G).

\textbf{Coverage and reproducibility.} Five families and nine models; newer, unreleased, and open-weight models are not included (the method is coverage-independent). A served model behind a name may change over time; we pin model version and collection date, but full reproducibility is not attainable for closed models --- what we measure is the \emph{deployed} model at a dated snapshot, itself the object of interest.

\textbf{Sampling and item-population uncertainty.} $N=40$ per model-item yields a standard error $\approx 0.08$ at $P=0.5$ (95\% half-width $\approx 0.16$); flips \emph{away} from the boundary are robust, flips \emph{near} it are not, which is why we report robust-flip counts under a $\pm 0.1$ dead-band. The 18 dilemmas are a small draw from the space of value trade-offs; per-pair genuine shares would move under a different item set.

\textbf{Further scope bounds.} Extremity is measured in a zero-context condition, so the reported base is the high-concentration endpoint of a preamble-sensitivity curve, not a neutral ground truth. The dilemmas are benign workplace value trade-offs with partial Schwartz/MFT tagging; harder dilemmas are future work. The second-domain (aesthetic) probe of Appendix~E is a single two-model, eight-item illustration and is not evidence of domain-generality. We measure what a model \emph{reports} it would choose under forced binary choice, not how it \emph{behaves} in situ; measuring that say--do gap is a natural extension, and a value--action divergence has been documented at the behavioral level \citep{huang2026knowing}. The dilemmas are benign and did not trigger provider use-policy responses; we do not treat any content-policy behavior as a variable.

\section*{Appendix J. A Reasoning-Effort Sweep Moves Engagement but Not Concentration or Lock}

The identifiability failure reported in the main paper is measured under a bare single-letter
forced choice, which invites the objection that the models were simply not given enough room to
reason. This appendix tests that objection directly on one model by sweeping the reasoning-effort
setting across its full range and asking whether the choice distribution follows.

\textbf{Model and condition.} The sweep uses \texttt{claude-opus-5}, a \emph{later and different}
model from the \texttt{claude-opus-4-8} measured elsewhere in this paper; we refer to it only by
its full identifier to keep the two apart. All 18 dilemmas, $N=40$ repetitions per dilemma per
effort level, position counterbalanced by repetition index, English prompt wording (Appendix~F),
$\texttt{max\_tokens}=8192$. The \texttt{thinking} parameter is left unset, which on this model
means adaptive thinking is active by default. This is the deployment-realistic case of a
practitioner who sets an effort level and does not touch the thinking configuration.

\textbf{The manipulation lands.} Table~J1 shows that raising the effort level
monotonically raises the rate at which the model actually spends reasoning tokens, from none at
all at the two lowest levels to roughly a fifth of prompts at the highest; the intervals at the
extremes do not overlap. Effort is therefore doing something measurable on this prompt.

\textbf{The outcome does not move --- but the pooled rate is the wrong place to look for it.}
Table~J2 shows the pooled forced-choice distribution over the same cells. It
sits between 47.9\% and 51.0\% option-A at every level and every interval overlaps every other. By
this paper's own argument that number cannot carry the conclusion: a pooled rate near 0.5 is
exactly the signature this paper shows to be non-identifiable. The two by-position columns in that table differ by design rather than by bias: a content-tracking model answers $v_1$ under one orientation and $1-v_1$ under the other, so counterbalancing is expected to split them. The split is consistent with the item-level concentration reported alongside it. We therefore report the two
per-item statistics the main text relies on, computed on the same 3{,}600 draws:

\begin{center}
\captionsetup{type=table}
\caption*{\textbf{Table J3.} Reasoning engagement and choice concentration at each effort level (\texttt{claude-opus-5}, English items, $N=40$ per model-item).}

\small
\setlength{\tabcolsep}{5pt}
\begin{tabular}{lccc}
\toprule
effort & reasoning engaged & extremity & position-lock \\
\midrule
low    & 0 of 720   & 0.947 & 0.056 \ (1/18) \\
medium & 0 of 720   & 0.961 & 0.000 \ (0/18) \\
high   & 24 of 720  & 0.956 & 0.056 \ (1/18) \\
xhigh  & 81 of 720  & 0.925 & 0.000 \ (0/18) \\
max    & 157 of 720 & 0.936 & 0.000 \ (0/18) \\
\bottomrule
\end{tabular}
\end{center}

Neither moves: extremity stays in 0.925--0.961 and position-lock stays at or below 0.056 (at most
one item of eighteen) across the whole range. The pooled rate near 0.5 is not indifference here;
it is per-item concentration of about 0.94 cancelling across items that commit in opposite
directions --- the aggregation artifact this paper is about, appearing in our own probe. Reading
only the pooled column would have told us nothing either way.

\textbf{What this licenses, and what it does not.} It licenses one inference: the position-linked
identifiability failure reported here is not an artifact of an under-provisioned reasoning budget
\emph{in this sweep}: raising the effort level moved reasoning engagement from 0 to 21.8\% and
moved neither per-item concentration nor position-lock, so a reviewer need not read the main
result as ``the models would have separated if you had let them think.'' The inference is bounded by the sweep --- one model, one prompt language, one
thinking configuration --- and we do not extend it to the identifiability failure in general.
It does not license the converse claim that reasoning is irrelevant to value expression in
general. The manipulation here raises the \emph{probability} that the model reasons, not the
depth of reasoning conditional on reasoning, and one model at one prompt wording is not the
population. It also does not transfer directly to the numbers in \S5. Those were collected in Korean with
the thinking parameter set explicitly, and Appendix~L includes a \texttt{claude-opus-5} cell
--- the same model swept here --- which engaged reasoning on 492 of 720 draws (68.3\%) against the
21.8\% recorded here at maximum effort. The two differ in prompt language and in whether
\texttt{thinking} was sent explicitly, and they differ by a factor of three. We report both rather
than reconciling them: that a single model's reasoning-engagement rate moves this much between two
nominally maximal-effort conditions is itself an instance of the access-conditioning this paper
documents. It also means the direction of the language effect is not constant across models ---
for \texttt{claude-opus-4-8} English elicited more engagement, here Korean does --- so we make no
claim about which language elicits more reasoning.

\textbf{Collection note.} An API credit exhaustion truncated the first pass at 2{,}832 of the
intended 3{,}600 calls; the 768 affected cells were re-collected under the identical configuration
and merged back, leaving every cell at its full $N=40$ per item. The merge replaced only the failed
records (all 2{,}832 originally successful records are byte-identical before and after), and
the re-collected rows are flagged as such in the file included in the code and data supplement. Of the 3{,}600 calls, every one
parsed to a valid letter and every one terminated normally.


\begin{table}[t]\centering\small
\caption*{\textbf{Table J1.} Manipulation check: reasoning-effort level versus the rate at which visible thinking is engaged. Counts are reported with their denominators; intervals are Wilson 95\%.}
\label{tab:effort-manip}
\begin{tabular}{lrrl}\toprule
effort & thinking engaged & rate & 95\% CI \\ \midrule
low & 0/720 & 0.0\% & [0.0, 0.5] \\
medium & 0/720 & 0.0\% & [0.0, 0.5] \\
high & 24/720 & 3.3\% & [2.3, 4.9] \\
xhigh & 81/720 & 11.2\% & [9.1, 13.8] \\
max & 157/720 & 21.8\% & [18.9, 25.0] \\
\bottomrule\end{tabular}\end{table}

\begin{table}[t]\centering\small
\caption*{\textbf{Table J2.} Outcome: forced-choice distribution by reasoning-effort level, with position counterbalancing. Every level's interval overlaps every other's, so these data do not show an effect of effort on the choice distribution.}
\label{tab:effort-outcome}
\setlength{\tabcolsep}{4pt}
\begin{tabular}{lrll}\toprule
effort & option A & rate (95\% CI) & by position \\ \midrule
low & 345/720 & 47.9\% [44.3, 51.6] & 123/360, 222/360 \\
medium & 348/720 & 48.3\% [44.7, 52.0] & 113/360, 235/360 \\
high & 354/720 & 49.2\% [45.5, 52.8] & 125/360, 229/360 \\
xhigh & 367/720 & 51.0\% [47.3, 54.6] & 133/360, 234/360 \\
max & 355/720 & 49.3\% [45.7, 53.0] & 132/360, 223/360 \\
\bottomrule\end{tabular}\end{table}


\section*{Appendix K. Does the Lock Persist on Later Models?}

The nine-model set is a dated snapshot (Appendix~G), so a reader may reasonably ask whether the
position-lock we report is an artifact of the particular model versions we froze. We can give a
partial answer without a new collection. The tier-by-generation probe of \S5 recorded the answer
letter as well as the reasoning-engagement rate, so the same position-lock and extremity
statistics can be computed on it. Four Anthropic models were collected there in one batch under
one protocol --- Korean prompts, adaptive thinking at maximum effort, $N=40$ per item over the
same 18 dilemmas, \texttt{max\_tokens = 8000}. This is a strongly reasoning-permitting,
generation-matched batch; it is not the same budget as Table~1 of the main paper, which used 16000.

\begin{center}
\captionsetup{type=table}
\caption*{\textbf{Table K1.} Extremity and position-lock for four models in the generation-matched batch (\texttt{max\_tokens = 8000}, $N=40$ per model-item).}

\begin{tabular}{lcc}
\toprule
model & extremity & position-lock \\
\midrule
\texttt{claude-opus-4-8}   & 0.72 & 0.22 \ (4/18) \\
\texttt{claude-opus-5}     & 0.83 & 0.17 \ (3/18) \\
\texttt{claude-sonnet-4-6} & 0.78 & 0.22 \ (4/18) \\
\texttt{claude-sonnet-5}   & 0.67 & 0.22 \ (4/18) \\
\bottomrule
\end{tabular}
\end{center}

Two things follow, and a third does not. First, the lock does not disappear on the later models:
every model here still answers with the same letter under both orientations on three or four of
the eighteen items, in the condition most favorable to reasoning. Second, extremity varies
across these four models (0.67--0.83) while position-lock is nearly flat (0.17--0.22), which is
consistent with the paper's argument that the lock fraction is the more stable of the two
quantities. What does \emph{not} follow is that the newer models lock less: the gap between
0.22 and 0.17 is a single item out of eighteen, and we do not read it as a difference. For the same
reason we flag rather than explain one discrepancy: \texttt{claude-sonnet-5} appears at
position-lock 3/18 in Table~1 of the main paper and 4/18 here, at identical extremity 0.67, across two batches with
different reasoning budgets. One item is within what this measurement resolves at $N=40$, and we
report both numbers with their batch rather than picking one.

We deliberately do not merge these numbers into Table~1 of the main paper. That table reports a three-condition
progression collected on 2026-07-17 at \texttt{max\_tokens = 16000}; the batch above used 8000,
and a paper whose thesis is that the reasoning budget conditions the measurement cannot treat two
different budgets as one condition. For the same reason we do not compare
\texttt{claude-opus-4-8}'s 0.72 here against its 0.64 in Table~1 of the main paper. The comparison that is clean is
the one within this batch, which is the one we make.

\section*{Appendix L. Reasoning Engagement by Tier and Generation}

At maximum effort, reasoning was engaged on 127 of 720 \texttt{claude-opus-4-8} prompts (17.6\%) and on 598 of 720 \texttt{claude-sonnet-5} prompts (83.1\%). We report the two separately rather than pooled. Engagement is also strongly item-dependent: for \texttt{claude-opus-4-8} it ranges from 32 of 40 draws on one dilemma to exactly 0 of 40 on nine of the eighteen, so no single rate characterizes the item set. To see whether that gap is attributable to the difference in model line or to the difference in generation, we measured the two remaining cells under the identical protocol: \texttt{claude-opus-5} engaged on 492 of 720 (68.3\%) and \texttt{claude-sonnet-4-6} on 711 of 720 (98.8\%). Within these four models the gap between the two earlier ones is large and the gap between the two later ones is small. The direction of change across generations is opposite in the two lines (Opus rising, Sonnet falling). We therefore state this as an observation about these four models and do not generalize it to a property of model tiers or of model generations; all four are from a single provider.

\nocite{zhang2026moralcomposition}
\nocite{zhang2026moralcomposition}
\bibliography{refs}
\end{document}